\documentclass{article}

\usepackage[preprint]{neurips_2026}

\usepackage[utf8]{inputenc}
\usepackage[T1]{fontenc}
\usepackage[hidelinks]{hyperref}
\usepackage{url}
\usepackage{booktabs}
\usepackage{amsfonts}
\usepackage{nicefrac}
\usepackage{microtype}
\usepackage{graphicx}
\usepackage{amsmath,amssymb}
\usepackage{xspace}

\hypersetup{
  pdftitle={How Reproducible Are Evaluation Conclusions? A Self-Audit of LLM-Inferred Prompt Structure},
  pdfauthor={Dipankar Sarkar}
}

\title{How Reproducible Are Evaluation Conclusions?\\ A Self-Audit of LLM-Inferred Prompt Structure}

\author{%
  Dipankar Sarkar\\
  Skelf Research\\
  \texttt{dipankar@skelfresearch.com}\\
  \url{https://skelfresearch.com}
}

\begin{document}

\maketitle

\begin{abstract}
Evaluations of LLM systems routinely average over small prompt sets and report models as a ranked table. We ask how much confidence such a table deserves, using LLM-based prompt-structure inference as the case study: eight open model variants across five families and 8B to 675B parameters, caching disabled, 293 raw intermediate representations persisted.

The measured phenomenon is unstable to begin with. Identical calls do not reliably recover identical structure, with mean node-set Jaccard from 0.39 to 0.96 and 72\% of prompt-model cells never node-set-perfect.

Auditing the evaluation weakens its conclusions further, and this is our main contribution. Under a joint cluster bootstrap over prompts, only the \emph{bottom} of the ranking is firm: the two least reproducible models hold rank in 99\% and 86\% of replicates, the middle four in 27\% to 48\%, and the top two in 68\% each, so the table identifies the worst model reliably but does not reliably identify the best. Two equally defensible rules for merging repeated campaigns change four of eight rows and move the study-wide headline by 7 percentage points. Checking the inferred structure against ground-truth annotations shows reproducibility cannot be read as accuracy. And four of the eight endpoints were withdrawn within ten weeks of measurement, so the study as specified can no longer be run.

Small-sample LLM evaluations can therefore look far more definitive than their evidence supports. We recommend reporting rank stability, per-cell provenance, executed sensitivity comparisons, raw per-run outputs, and a measurement date alongside any ranking.
\end{abstract}

\section{Introduction}
\label{sec:intro}

A wave of recent systems treats prompts as programs: as queries to be compiled~\citep{beurerkellner2023lmql}, as pipelines to be optimised~\citep{khattab2024dspy}, or as artefacts over which to define test adequacy~\citep{tambon2026promptcoverage,tan2026skillcoverage}. Applied to a prompt not authored in the system's own format, all of them face the same prerequisite: recover the prompt's structure. The natural way to do that is to ask a language model. The natural worry is that the model may not answer the same way twice.

That worry is well founded and well documented in general terms: temperature-zero decoding does not guarantee determinism~\citep{ouyang2025nondeterminism,atil2025nondeterminism}, and repeated-run consistency has been studied for structured extraction~\citep{yao2026consistency}. Our first contribution measures it specifically for prompt-structure inference, across model families and parameter scales, with the raw outputs released.

Our central contribution is a claim-level audit of that evaluation. Having ranked eight model variants by reproducibility, we subjected that ranking to the same scrutiny we had applied to the models, and the full ranking does not survive. Four further results depended on undocumented analysis choices: the precedence rule for prompts measured in more than one campaign, absolute versus relative stabilisation tolerance, a criterion satisfied by construction at its endpoint, and the timeout at which a slow model is classified unavailable. Each is an analysis, specification, or instrumentation choice capable of moving a headline without changing the underlying model outputs.

Table~\ref{tab:audit} collects seven claims and assumptions examined in the self-audit and records which survive. The failure mode is likely to matter beyond this study. Evaluations built from tens of prompts and a handful of models, with a metric averaged per model and sorted, are cheap to run and easy to cite, and the sorting is the part a reader retains and the part the sample size supports least.

\paragraph{Structure.}
The paper proceeds in four steps, from the instability of the measurement to the instability of everything built on top of it.

\begin{description}\itemsep1pt
  \item[\S\ref{sec:results}. Does the measurement reproduce?] Jaccard and ID-stability across repeated identical calls; whether the gap is a property of the model or of scale within a family; whether corpus type shifts it; and what drifts when the IR drifts.
  \item[\S\ref{sec:rq6}. Do the conclusions reproduce?] How many reruns a consumer must average before its signal stabilises, and whether the resulting ranking survives resampling and defensible changes of analysis.
  \item[\S\ref{sec:correctness}. Reproducibility is not validity.] Whether a stable IR is also a correct one, against ground-truth annotations.
  \item[\S\ref{sec:rq7}. Reproducibility over time.] Whether the study can be run again at all, ten weeks later.
\end{description}

\section{Background}
\label{sec:bg}

\paragraph{The inference task.}
Most prompts in the wild are unstructured natural language. To apply a program-style toolchain, a system must first ask a model to read the prompt and emit a serialisable intermediate representation (IR): a list of typed nodes (\texttt{Rule}, \texttt{Condition}, \texttt{Breakpoint}, \texttt{Directive}), each carrying a string id, a category, an optional priority, and a source range. We treat this IR as a black-box output and ask only whether the same prompt yields the same one.

\paragraph{A concrete input--output example.}
One evaluated support-ticket prompt contains the line
\texttt{// @rule RULE:urgency\_assessment category=policy priority=2}, followed by the instruction to assess a ticket as high, medium, or low urgency. One run emitted a node with id \texttt{RULE:urgency\_assessment}, kind \texttt{Rule}, \texttt{inferred=false}, category \texttt{policy}, and priority 2. The same run also added an unmarked \texttt{RULE:conditional\_logic} section with \texttt{inferred=true}. Thus one IR can contain both copied explicit structure and structure recovered from ordinary prompt text; the persisted JSON records also include labels, triggers, confidence, and source ranges.

\paragraph{Two regimes, and which one we mostly measure.}
The inference instruction tells the model that explicit \texttt{// @rule} annotations may be present and should be emitted with \texttt{inferred=false}, and that further rules be inferred and marked \texttt{inferred=true}. The task therefore has two components, transcribing marked structure and recovering unmarked structure, and our synthesised corpus is annotated, so it is dominated by the first: across the 293 persisted IRs the share of nodes marked explicit runs from 64\% to 100\%, most models between 77\% and 92\%. Table~\ref{tab:rq1} is thus mostly \emph{transcription} reproducibility, which sharpens the headline rather than softening it: these models are shown delimited structure, told to copy it, and still do not return the same node set twice. The unannotated regime is the recovery campaign of \S\ref{sec:results}.

\paragraph{Why reproducibility is the precondition.}
A linter, a coverage tool, or an agent that debugs a teammate's prompt cannot offer the guarantee that ordinary static analysis offers routinely, that the same input yields the same diagnosis tomorrow, if the structure it reasons over is itself resampled on every call. We are measuring that precondition, not the downstream tool.

\section{Method}
\label{sec:method}

\paragraph{Models.}
We used one commercial open-model endpoint and attempted ten models spanning family, parameter scale, and reasoning style. Eight met our inclusion threshold of at least 3 prompts completing at least 2 of $R$ runs: \texttt{gemma3:12b} (Google); \texttt{gpt-oss:20b} and \texttt{gpt-oss:120b} (OpenAI-style); \texttt{ministral-3:8b} and \texttt{mistral-large-3:675b} (Mistral); \texttt{minimax-m2.1} and \texttt{minimax-m2.7} (MiniMax); \texttt{qwen3-coder-next} (Alibaba).

Two did not. \texttt{deepseek-v3.1:671b} returned no successful inference across the attempts we logged; \S\ref{sec:rq7} makes our client timeout the leading candidate, but the model is retired and we cannot now establish the cause. \texttt{glm-4.7} returned 5 successful runs across 2 prompts (Jaccard 0.80 and 1.00), real data falling below our 3-prompt inclusion threshold. It is excluded from Table~\ref{tab:rq1} by that threshold rather than by failure, a distinction worth recording since the artefact contains those 5 IRs.

\paragraph{Corpus.}
Two corpora. The synthesised corpus on disk holds 83 annotated prompts: 80 generated across a rotating list of 30 application domains and three hand-authored seeds. The generator's released instruction requests at least four rules, two conditions, and nested branching, and the generated files were retained rather than paraphrased into additional variants. The campaigns reported here measured 22 distinct prompt files, and raw IRs were persisted for 20 (the flag that saves them was added partway through). Repeated runs submit the same prompt file and inference instruction with caching disabled; there are no hidden prompt variants. We state the counts 83, 22, and 20 because they are easy to conflate. Table~\ref{tab:rq1} and the drift analysis therefore rest on 22 and 20 prompts respectively, not on 83.

The second corpus is 40 license-gated real-world \texttt{SKILL.md} prompts collected from 40 distinct public GitHub repositories under permissive licences (38 MIT, 1 Apache-2.0, 1 Unlicense), with attribution and content-hash deduplication; 23 completed at least 2 runs within budget. Real prompts are passed to inference as-is, which is the setting a tool ingesting existing prompts actually faces.

\paragraph{Metrics.}
For a prompt we run inference $R$ times ($R$ between 2 and 5, set per campaign by rate-limit headroom) with caching disabled, obtaining IRs $\{I_1,\dots,I_R\}$.
Write $\mathrm{ids}(I)$ for an IR's node-id set. \emph{Node-set Jaccard} averages agreement over run \emph{pairs}: the mean of $|\mathrm{ids}(I_i)\cap\mathrm{ids}(I_j)|/|\mathrm{ids}(I_i)\cup\mathrm{ids}(I_j)|$ over all $\binom{R}{2}$ of them. \emph{ID-stability} instead demands agreement across \emph{all} $R$ runs at once: $|\bigcap_i \mathrm{ids}(I_i)|/|\bigcup_i \mathrm{ids}(I_i)|$. The first is the more forgiving of the two.
A cell is \emph{node-set-perfect} when both equal 1. We name it that way deliberately: it means the node-id sets agreed, not that the IRs were identical. Metadata and hierarchy can still differ, and in our data they sometimes do, so calling this "perfect reproducibility" would overstate what the metric sees.

\paragraph{Drift decomposition.}
When Jaccard $<1$ we classify each pairwise disagreement as exactly one of: \emph{id-label drift} (same node count and kind distribution, different ids); \emph{count drift} (different node counts or kind distribution); \emph{metadata drift} (same ids and kinds, differing source range, category or other per-node metadata). Calling the third category \emph{hierarchy} drift would overstate it: the IR as emitted is a flat node list with no parent-child relation, so the bucket measures per-node metadata disagreement, chiefly source ranges.

\paragraph{Statistics.}
Means carry 95\% percentile bootstrap intervals over 10{,}000 prompt resamples. Within-family contrasts use a paired Wilcoxon signed-rank test on per-prompt Jaccard, restricted to prompts both models completed. Because tie handling changes the result at these sample sizes, we report both the zero-splitting and the zero-dropping convention. Rank stability is estimated by a joint cluster bootstrap: one prompt multiset is drawn per replicate, every model is scored on that same draw, and the table is re-sorted.

The main table is an incomplete, unbalanced panel rather than a fully crossed benchmark: campaigns covered different prompt subsets, and completion failures further changed each row's support. Raw cross-row differences are therefore descriptive, not controlled pairwise effects. The within-family tests above are the only model contrasts we treat as paired, and they use only prompts completed by both variants; Table~\ref{tab:rq1} reports each row's actual $n$ so the unmatched coverage remains visible.

\paragraph{Merge precedence.}
Measurements were collected in several campaigns at different wall-clock budgets, so a given prompt-model pair can appear more than once. We resolve duplicates by preferring the higher-budget campaign, and we publish the per-cell provenance table alongside the results. This rule is stated because, as \S\ref{sec:rq6} shows, it is not neutral.

\section{Does the measurement reproduce?}
\label{sec:results}

The first four questions ask whether the phenomenon is stable enough to support a ranking at all.

\subsection{How reproducible is it?}

Table~\ref{tab:rq1} gives the picture. Mean Jaccard runs from 0.39 to 0.96 and ID-stability from 0.22 to 0.96. Across all 127 prompt-model cells, 35 (28\%) are node-set-perfect.

The bootstrap intervals deserve as much attention as the point estimates. \texttt{gpt-oss:120b} at 0.39 has a 95\% interval of $[0.25, 0.54]$, and \texttt{minimax-m2.7} at 0.70 has $[0.47, 0.89]$ on $n=6$. These are not narrow measurements. Most adjacent pairs have overlapping intervals; the exceptions are the well-separated top and bottom of the table, where for instance \texttt{gemma3:12b} at $[0.90,0.98]$ and \texttt{mistral-large-3:675b} at $[0.73,0.86]$ do not overlap.

\begin{table}[t]
\caption{Stability across eight model variants on the synthesised corpus, sorted by Jaccard. $n$ is prompts completing at least 2 of $R$ runs. Brackets are 95\% percentile bootstrap intervals over 10{,}000 prompt resamples. "Rank held" is the fraction of joint-cluster-bootstrap replicates in which the model keeps the position shown here; prompts are resampled once per replicate and all models scored on the same draw, so the pairing is preserved.}
\label{tab:rq1}
\centering
\small
\setlength{\tabcolsep}{4pt}
\begin{tabular}{llrccrr}
\toprule
Model & Family & $n$ & Jaccard [95\% CI] & ID-stab [95\% CI] & Perfect & Rank held \\
\midrule
\texttt{minimax-m2.1}    & MiniMax      &  3 & 0.96 [0.89, 1.00] & 0.96 [0.89, 1.00] &  2/3  & 68\% \\
\texttt{gemma3:12b}      & Google       & 22 & 0.94 [0.90, 0.98] & 0.92 [0.85, 0.98] & 15/22 & 68\% \\
\texttt{mistral-large-3:675b} & Mistral & 19 & 0.80 [0.73, 0.86] & 0.71 [0.62, 0.80] &  3/19 & 48\% \\
\texttt{ministral-3:8b}  & Mistral      & 19 & 0.77 [0.66, 0.87] & 0.71 [0.57, 0.83] &  6/19 & 27\% \\
\texttt{gpt-oss:20b}     & OpenAI-style & 20 & 0.76 [0.65, 0.85] & 0.69 [0.56, 0.80] &  4/20 & 47\% \\
\texttt{minimax-m2.7}    & MiniMax      &  6 & 0.70 [0.47, 0.89] & 0.68 [0.44, 0.89] &  2/6  & 48\% \\
\texttt{qwen3-coder-next} & Alibaba     & 20 & 0.58 [0.49, 0.66] & 0.47 [0.37, 0.57] &  0/20 & 86\% \\
\texttt{gpt-oss:120b}    & OpenAI-style & 18 & 0.39 [0.25, 0.54] & 0.22 [0.06, 0.42] &  3/18 & 99\% \\
\midrule
\multicolumn{7}{l}{\textit{Below inclusion threshold:} \texttt{glm-4.7}, 5 successful runs over 2 prompts;} \\
\multicolumn{7}{l}{\texttt{deepseek-v3.1:671b}, 0 successful runs over the 27 attempts in the archived log.} \\
\bottomrule
\end{tabular}
\end{table}

Perfect reproducibility is largely a function of which model is chosen rather than which prompt: \texttt{gemma3:12b} is node-set-perfect on 68\% of its prompts, while \texttt{qwen3-coder-next} never is on any of its 20.

\subsection{Model dependence, and scale within a family}

The spread across model means is 0.58, with a standard deviation of 0.18. That much is robust.

The within-family scaling claim is not. We can test three families:

\begin{itemize}
  \item \textbf{OpenAI-style.} On the 18 prompts both completed, \texttt{gpt-oss:120b} scores 0.386 against \texttt{gpt-oss:20b}'s 0.766, a paired difference of $-0.380$ (Wilcoxon $p=0.0046$ splitting zeros, $p=0.0036$ dropping them; matched-pairs rank-biserial $-0.80$). The larger model is markedly less reproducible.
  \item \textbf{Mistral.} On 18 common prompts the larger model is \emph{better} by $+0.061$, and the difference is not significant ($p=0.37$).
  \item \textbf{MiniMax.} Only 2 prompts are common to both variants, and both score 1.000 on each. This comparison carries no information, and the apparent 0.96-versus-0.70 gap in Table~\ref{tab:rq1} is driven entirely by non-overlapping prompts and differing completion rates.
\end{itemize}

One family shows strong inverse scaling, one shows weak positive scaling, and one shows nothing. The defensible claim is the existential one: scaling a family up can substantially worsen structural reproducibility, and \texttt{gpt-oss} is a clear instance. The general claim, that bigger is not better, is not supported by three families of which one is informative.

\subsection{Transcription versus recovery}

The panel above is mostly transcription (\S\ref{sec:bg}). To measure recovery we ran a second campaign, crossed by design: 11 currently-served models on the same 19 genuinely unannotated real \texttt{SKILL.md} files, three runs each, one time window. The design calls for 627 outputs and 557 persisted, so the analysed panel is not fully crossed: we analyse the 196 cells with at least two runs, and the 70 missing outputs are not missing at random, since a model that fails is likelier to fail again. Appendix~\ref{app:rq5} has the per-cell completion counts.

Recovery scores far lower, though the comparison is not clean: annotation status, corpus, prompt content, model panel and completion pattern all change together between the two regimes, so what follows is a level difference and not an isolated effect of recovery. Mean node-set Jaccard over the 196 cells is \textbf{0.22}, against 0.39 to 0.96 per model in the annotated regime, and 30 cells (15\%) score exactly zero, repeated runs of the same model on the same prompt sharing no node id at all. Per model the exact-id mean runs from 0.73 (\texttt{gemma4:31b}) down to 0.05 (\texttt{nemotron-3-nano:30b}); under the normalised metric introduced below the ordering at the bottom changes slightly, with \texttt{minimax-m2.7} lowest at 0.11.

\paragraph{The disagreement is substantive, not nominal.}
An exact-id metric is unfair here in a way it is not on annotated input, since there is no canonical naming: two runs can agree completely about content yet score zero by calling one unit \texttt{SECTION\_WHEN\_TO\_USE} and \texttt{RULE:WHEN\_TO\_USE}. We therefore also report Jaccard after stripping prefixes and normalising separators, and agreement on node count, defined as $1 - |n_i - n_j| / \max(n_i, n_j)$ and so a graded score rather than a test of equality. Normalisation lifts the mean only from 0.22 to \textbf{0.25}. Count agreement is higher at 0.70, so models are closer on \emph{how many} units exist than on \emph{which}, though exact equality of node count is far rarer than that 0.70 suggests, holding in only 100 of 508 run pairs (20\%). Naming conventions account for about three points of the gap. We stop short of calling the remainder genuine structural disagreement: the normaliser is ASCII-only, mapping 152 ids to the empty string and costing 171 within-run distinct ids to collisions across 34 runs, and it does not address synonyms. A Unicode-preserving recomputation still rounds to 0.25, so the headline is robust; what the residue \emph{means} is not.

\paragraph{What we cannot conclude.}
Only four models appear in both regimes, since half the original panel has since been retired (\S\ref{sec:rq7}). On those four the ordering is broadly preserved (Spearman $\rho=0.80$) but at $n=4$ this is not significant ($p=0.20$), and two of the four change rank. We had expected the ranking to scramble between regimes and it does not visibly do so; we simply cannot test it at this sample size. What the data support is the level difference, not a claim about transfer.

For continuity with the earlier campaign: on the real corpus under the original protocol \texttt{gemma3:12b} attained 0.85 with 13 of 23 node-set-perfect, against 0.94 on the synthesised corpus, a drop of 0.10. That model can no longer be queried.

\subsection{What drifts}

Across 98 prompt-model cells with persisted raw IRs, drift is dominated by count disagreements: 19\% id-label, 49\% count, 31\% metadata, as a macro mean over cells. The ordering is stable across every aggregation we tried (pooling raw disagreement pairs gives 19/47/33; restricting to cells that actually disagree gives 20/49/31), so the headline is safe even though the exact figures are not.

Models do not mostly rename a stable skeleton, which id canonicalisation would fix; they disagree about how many nodes there are, so canonicalisation would address under a fifth of the drift. One panel mismatch: the 98 cells include \texttt{glm-4.7} and exclude \texttt{gpt-oss:20b}, whose campaign predated the flag that saves raw IRs, so the drift panel and Table~\ref{tab:rq1} are both eight models but not the same eight.

\section{Do the conclusions reproduce?}
\label{sec:rq6}

The rest of the paper turns the same scrutiny on the evaluation itself, asking here whether the answers above survive resampling and defensible changes of analysis.

\subsection{How many reruns a consumer needs}

Our stabilisation analysis was wrong twice, and both errors are analysis-specification failures rather than results about models, so we summarise here and give the detail in Appendix~\ref{app:rq5}.

The criterion is satisfied by construction at its endpoint, since with $R$ runs the long-run mean \emph{is} the mean of all $R$, so $r=R$ always passes and "did not stabilise" is unreachable, accounting for 48 of 98 cells. And the tolerance was documented as relative to the long-run mean but implemented as an absolute 0.05 \emph{nodes}; on trajectories whose long-run means range from 5.3 to 18.0 nodes those two readings give 80 and 50 of 98 cells stabilising. Nothing in the data distinguishes them.

We also do not measure a downstream tool: the signal is IR node count, a proxy for any quantity a consumer might derive, and no linter finding or coverage value is computed. All we will claim is that among cells stabilising strictly before their endpoint the median is $r=1$ under both readings, so where averaging helps it helps immediately. This design does not support a recommendation that any fixed number of reruns suffices.

\begin{table}[t]
\footnotesize
\setlength{\tabcolsep}{3pt}
\caption{Seven claims and assumptions examined in our self-audit. Only one remains strongly supported. Each row is established in the section named; none required new measurements.}
\label{tab:audit}
\begin{tabular}{@{}p{0.37\textwidth}p{0.24\textwidth}p{0.33\textwidth}@{}}
\toprule
Claim or assumption & Audit & Outcome \\
\midrule
\texttt{gpt-oss:120b} is the least reproducible model & joint cluster bootstrap (\S\ref{sec:rq6}) & \textbf{Holds.} Stays last in 99\% of replicates \\
\texttt{minimax-m2.1} is the most reproducible model & joint cluster bootstrap (\S\ref{sec:rq6}) & \textbf{Not supported.} Holds rank in 68\%; which of the top two leads is undetermined \\
The reported ranking is robust to campaign-merging choices & alternative merge rule (\S\ref{sec:rq6}) & \textbf{Sensitive.} Four of eight rows change; headline moves 7.1 points \\
Repeated inference stabilises after $r$ runs & tolerance specification (Appendix~\ref{app:tolerance}) & \textbf{Specification-dependent.} 80 vs 50 of 98 cells under two readings of one flag \\
DeepSeek and Zhipu are structurally inaccessible & harness audit (\S\ref{sec:rq7}) & \textbf{Withdrawn.} Instrument is the leading candidate; cause unresolved \\
A stable IR is a correct IR & ground-truth annotations (\S\ref{sec:correctness}) & \textbf{Not supported.} The two orderings diverge \\
The study can be rerun & endpoint audit (\S\ref{sec:rq7}) & \textbf{No.} Four of eight endpoints retired within ten weeks \\
\bottomrule
\end{tabular}
\end{table}

\subsection{Does the ranking survive resampling and analysis choices?}

\paragraph{The full ranking does not survive resampling.}
The final column of Table~\ref{tab:rq1} reports how often each model retains its listed rank under a joint cluster bootstrap: one prompt multiset is drawn per replicate and every model scored on that draw, preserving the pairing an independent per-model resample would destroy. Of 10{,}000 replicates, 9{,}583 were usable; the rest contained no prompt measured by the $n=3$ model.

Only the bottom holds: \texttt{gpt-oss:120b} stays last in 99\% of replicates and \texttt{qwen3-coder-next} seventh in 86\%. The middle four range from 27\% to 48\%, close to a coin flip. The top is no firmer: \texttt{minimax-m2.1} and \texttt{gemma3:12b} hold position in only 68\% each, and which leads is undetermined at $n=3$ against $n=22$.

So the table supports "\texttt{gpt-oss:120b} is the least reproducible model we tested" but does not reliably identify the most reproducible. Retention frequency is a bootstrap diagnostic, not a posterior probability, and it is strict: a model displaced by one position may still be well separated from the field. Rank confidence intervals~\citep{neuhof2026rankintervals} would be the more principled instrument, and we report retention only as a diagnostic.

We claim the application, not the method. Rank uncertainty for leaderboards has an emerging machinery: error bars on evaluations~\citep{miller2024errorbars}, rank confidence and prediction intervals~\citep{neuhof2026rankinguncertainty,neuhof2026rankintervals}, leaderboard tops flipped by sub-1\% perturbations~\citep{oyarhoseini2026leaderboardstability}, and naive standard errors 40 to 60\% too small once judge and phrasing variance are accounted for~\citep{messing2026hiddenerror}. We introduce no new rank-uncertainty estimator; our contribution is applying uncertainty analysis as one stage of an end-to-end audit of a concrete evaluation.

\paragraph{The table moves under a defensible change of analysis.}
Our measurements accumulated over several campaigns at different budgets, so some prompt-model pairs were measured more than once. Preferring the higher-budget campaign is one reasonable rule; preferring the earliest and topping up from later ones only for prompts not yet covered is another, which is what an analyst assembling results incrementally would do, and it was applied here without being recorded.

Recomputing the entire table under both rules, which \texttt{scripts/stats\_analysis.py} now does rather than asserting, changes the row of \emph{four} of the eight models (Appendix~\ref{app:merge}):

The study-wide headline moves by 7.1 percentage points. Three rows change rank: \texttt{minimax-m2.7} rises from sixth to fourth, \texttt{ministral-3:8b} falls from fourth to fifth and \texttt{gpt-oss:20b} from fifth to sixth, while \texttt{qwen3-coder-next} stays seventh. A sentence we would otherwise have written, that only one model is never node-set-perfect, becomes a different sentence under the other rule.

A first pass at this comparison reported agreement on seven of eight models and a four-point move, having substituted one row rather than applying either rule globally. Neither rule is wrong; not recording which was used, and not executing the comparison, is.

\paragraph{Sample sizes do not support the comparisons the format invites.}
Two rows of Table~\ref{tab:rq1} rest on $n=3$ and $n=6$. A table sorted by a point estimate invites comparison between every pair of rows, and we make no multiplicity correction for the 28 implied comparisons because the study was not designed to support them. At these sample sizes a sorted table is better read as a partition into "clearly reproducible", "clearly not", and "undetermined", and most of our models fall in the third group.

\section{Reproducibility is not validity}
\label{sec:correctness}

Agreement between runs is not agreement with truth. The synthesised prompts are annotated and the deterministic parser recovers those annotations exactly, so re-reading the 293 persisted IRs offline gives ground truth for the explicit structure with no new inference. Mean recall of the annotated node ids is 0.86, and mean precision \emph{restricted to nodes the model itself marked explicit} is 0.82: roughly one in five nodes a model asserted it had copied corresponds to no annotation in the file.

Stability and correctness also order the models differently, and how strongly they associate depends on which correctness metric is chosen, from Pearson $r=0.69$ under all-node precision to $r=0.90$ under explicit precision, with none of the four surviving the removal of the single low outlier. Appendix~\ref{app:correctness} has the per-model figures, the metric definitions and all four correlations; the panels differ too, stability being prompt-weighted over the merged cells and correctness run-weighted over 288 runs in 110 cells. We therefore claim only that the orderings diverge, and the practical consequence stands regardless: a reproducibility ranking cannot be read as an accuracy ranking, and a practitioner choosing on Table~\ref{tab:rq1} alone would take \texttt{mistral-large-3:675b} over \texttt{ministral-3:8b} and get the less accurate of the two.

\section{Reproducibility over time}
\label{sec:rq7}

We re-queried the endpoint for every model this study names. Of the eight variants in Table~\ref{tab:rq1}, \textbf{four are no longer served}, all withdrawn on 2026-07-15, a month after our measurements. The provider returns HTTP 410 with an explicit retirement date and reference id, so this is not a transient outage. Of the fifteen models originally named, ten are gone across two waves (2026-06-16 and 2026-07-15); Appendix~\ref{app:availability} has the audit.

The study as specified is therefore no longer executable. \texttt{gemma3:12b}, second in Table~\ref{tab:rq1} and the anchor of the annotated real-corpus comparison, cannot be queried by us or by a reviewer, so half the ranking \S\ref{sec:rq6} was careful to qualify now rests on subjects nobody can re-examine.

\paragraph{Re-examining the accessibility finding.}
An earlier version of this study called DeepSeek and Zhipu \emph{structurally} inaccessible. The likelier explanation is a single constant in our own client: the default analysis timeout returns 60{,}000\,ms, overridable by no flag, and anything slower is logged as a connection failure, which is the exact signature reported. Many current models emit a separate \texttt{reasoning} trace before answering, and our client caps no tokens, so a reasoning-heavy model runs past the deadline.

Probing 11 currently-served models twice on one prompt, under a tight output budget and a generous one (Appendix~\ref{app:attribution}), \textbf{eight fail under the tight budget and return valid JSON with room}; seven of those eight finish inside 60\,s, \texttt{nemotron-3-ultra} needs 77.7\,s and would be discarded whatever the budget. Of the three that work under both, \texttt{mistral-large-3:675b} and \texttt{gemma4:31b} emit no reasoning at all, while \texttt{glm-5.2} emits 4{,}372 characters, close to the median of 4{,}906 among the other eight. Reasoning volume therefore separates the extremes without cleanly partitioning the groups, and we do not claim it as the mechanism. Two hazards separate here: the timeout is the leading candidate for what broke this study, though the failure could not be reproduced, while the budget is an independent trap for anyone who sets a cap. Either makes reasoning-heavy families look unreachable, and DeepSeek and Zhipu are reasoning-model families.

We therefore withdraw the accessibility finding. The original evidence could not distinguish model, provider, timeout, schema or harness failure, and the five successful GLM runs already contradicted the family-level claim. We have not \emph{reproduced} the historical failure: those models are retired, our probe uses a simplified extraction prompt rather than the typed IR request, and the budget condition is not one the harness ever applied. The cause is unresolved; the instrument is the leading candidate.

\paragraph{Where this sits, and what it implies.}
Deprecation and silent revision are documented~\citep{ma2024promptworse,chen2024chatgptdrift}, and so is this failure, which we do not claim to be first to name. \citet{johnson2024deprecating} report that after OpenAI withdrew \texttt{text-davinci-003} the prompts from their own published study could no longer be resubmitted, leaving data and code inspectable but the generating process unrepeatable; \citet{angermeir2025reproducibility} find the same mechanism at venue scale, naming deprecated models the one reproduction blocker specific to commercial LLMs; and \citet{baltes2026guidelines} responds with guidelines whose fourth item is to report session traces. Our case adds a dated quantification on a study whose raw outputs survive: ten of fifteen named models gone between measurement (30 May to 11 June 2026) and audit (17 August 2026). It survives only because 293 raw IRs were persisted, from which the drift, stabilisation and correctness analyses regenerate offline; had we kept only summary CSVs, half this paper would be unfalsifiable. If an evaluation targets a hosted endpoint, treat raw per-run outputs as the primary artifact, date every measurement, and present the ranking as a dated document.

\section{Threats to validity}
\label{sec:threats}

\paragraph{Statistical instability and study defects are different failure modes.}
Run-to-run disagreement, bootstrap rank movement, and merge-rule sensitivity concern how much inference the sampled outputs support. The endpoint timeout and the two stabilisation mistakes instead concern our instrument and analysis specification. Those mistakes are evidence that an evaluation pipeline can fail under audit, not evidence of intrinsic LLM nondeterminism. We keep the two classes separate in the results and withdraw the accessibility claim whose instrument could not identify its cause.

\paragraph{One endpoint, one window, one instrument.}
Every measurement comes from a single commercial endpoint over a bounded period, so the per-model numbers describe a model-as-served, and describe it through an instrument whose 60-second timeout \S\ref{sec:rq7} shows is enough on its own to manufacture an availability finding. Nor can we prove which weights answered us: we queried by name and cannot verify the served artifact matched it, a gap \citet{zhang2026shadowapis} shows is not hypothetical, and silent revision under a stable name would be indistinguishable here from nondeterminism.

\paragraph{Mostly transcription, not recovery.}
Between 64\% and 100\% of nodes in the synthesised regime are explicit annotations the model was told to copy (\S\ref{sec:bg}). The recovery campaign runs eleven models on the unannotated corpus but only four appear in both panels, so the cross-regime ordering rests on those four, and annotation status, corpus, prompt content, model panel and completion pattern all change together.

\paragraph{Case-study external validity.}
Prompt-structure inference is a specialised task, and the study does not include a standard LLM benchmark or a downstream RAG, agent, or code-generation evaluation. A fully crossed second setting would be needed to estimate how often the same failures occur elsewhere. Our broader recommendations are therefore procedural safeguards motivated by this case, not an estimate of failure prevalence across LLM evaluation.

\paragraph{Narrow metrics, unequal $R$.}
Jaccard and ID-stability see node-id sets only; our drift decomposition finds metadata disagreement in cells scoring Jaccard 1, and the correctness check covers explicit annotations alone. $R$ varies from 2 to 5 and completion from 3/20 to 20/20, so node-set-perfection is not strictly commensurable across rows, and a prompt that repeatedly times out is plausibly one the model finds hard.

\section{Related work}
\label{sec:related}

Non-determinism under nominally deterministic settings is established: \citet{ouyang2025nondeterminism} document it for code generation at temperature zero, \citet{atil2025nondeterminism} propose determinism metrics, and \citet{yao2026consistency} argue run-to-run consistency be measured semantically rather than on surface form, which our drift decomposition supports from another direction. We claim no discovery that LLM outputs vary; our contribution is the task, the cross-family design, and the released raw IRs.

Systems that treat prompts as programs~\citep{beurerkellner2023lmql,khattab2024dspy} assume the program is authored in their own format, and recent adequacy and mutation work over prompts~\citep{tambon2026promptcoverage,tan2026skillcoverage,wei2024mile} operates on requirement semantics or in-context demonstrations. None performs the inference we study. On evaluation robustness, format~\citep{sclar2024sensitivity} and demonstration-order~\citep{lu2022fantastically} sensitivity show conclusions move when inputs are perturbed; our audit adds that they move when nothing is perturbed except which resample one drew.

\section{Conclusion}

Asking a language model for the structure of a prompt does not reliably give the same answer twice, even when that structure is marked and the model is told to copy it: Jaccard spans 0.39 to 0.96 and 72\% of cells are not node-set-perfect. The more consequential finding is about our own table: of the seven claims and assumptions we audit, one remains strongly supported under the audits applied (Table~\ref{tab:audit}), and every failure was found without new data.

Our recommendation is procedural: publish per-cell provenance, execute sensitivity comparisons rather than asserting them, state the merge rule and every tolerance's units, report rank stability beside any ranking, and date the leaderboard. The corpus, the 293 raw IRs and every script are released.

\appendix

\section{Model availability audit}
\label{app:availability}
Re-queried 2026-08-17 via \texttt{scripts/model\_availability.py}; raw output in \texttt{results/model\_availability.csv}.

\begin{table}[!h]
\caption{Availability of every model named in this study, re-queried on 2026-08-17. Two bulk retirement waves are visible. Of the eight variants in Table~\ref{tab:rq1}, four can no longer be queried; of the fifteen models named anywhere in the study, ten cannot.}
\label{tab:availability}
\centering
\small
\begin{tabular}{llc}
\toprule
Model & Role in the study & Status on 2026-08-17 \\
\midrule
\texttt{gpt-oss:20b}          & Table~\ref{tab:rq1} & available \\
\texttt{gpt-oss:120b}         & Table~\ref{tab:rq1} & available \\
\texttt{mistral-large-3:675b} & Table~\ref{tab:rq1} & available \\
\texttt{minimax-m2.7}         & Table~\ref{tab:rq1} & available \\
\texttt{gemma3:12b}           & Table~\ref{tab:rq1} & retired 2026-07-15 \\
\texttt{ministral-3:8b}       & Table~\ref{tab:rq1} & retired 2026-07-15 \\
\texttt{minimax-m2.1}         & Table~\ref{tab:rq1} & retired 2026-07-15 \\
\texttt{qwen3-coder-next}     & Table~\ref{tab:rq1} & retired 2026-07-15 \\
\midrule
\texttt{deepseek-v3.1:671b}   & attempted, 0 successful runs & retired 2026-07-15 \\
\texttt{glm-4.7}              & attempted, below threshold   & retired 2026-07-15 \\
\texttt{deepseek-v3.2}        & probed as substitute & retired 2026-07-15 \\
\texttt{glm-4.6}              & probed as substitute & retired 2026-06-16 \\
\texttt{qwen3-next:80b}       & probed as substitute & retired 2026-06-16 \\
\texttt{kimi-k2-thinking}     & probed as substitute & retired 2026-06-16 \\
\texttt{nemotron-3-super}     & probed as substitute & available \\
\bottomrule
\end{tabular}
\end{table}

\section{IR correctness against ground-truth annotations}
\label{app:correctness}

Precision is reported over the nodes the model marked explicit, where a mismatch is unambiguously an error. F1 is computed over \emph{all} inferred nodes and therefore penalises the implicit rules the instruction asks for and our ground truth cannot score; it is not the harmonic mean of the two adjacent columns and should not be read as one. Recall and explicit precision are the interpretable columns.

Correlation between per-model stability (\texttt{mean\_jaccard}) and each correctness metric, over the seven models measured both ways. The last pair drops \texttt{gpt-oss:120b}, the low outlier that carries most of the linear signal. No metric survives that drop at $\alpha=0.05$, which is why we claim only that the orderings diverge. Regenerated by \texttt{scripts/stability\_vs\_correctness.py}.

\begin{center}
\small
\begin{tabular}{lrrrrrr}
\toprule
Correctness metric & Pearson $r$ & $p$ & Spearman $\rho$ & $p$ & drop-one $r$ & $p$ \\
\midrule
Recall & $+0.76$ & 0.046 & $+0.50$ & 0.253 & $+0.28$ & 0.594 \\
Precision (all nodes) & $+0.69$ & 0.085 & $+0.75$ & 0.052 & $+0.58$ & 0.229 \\
Precision (explicit) & $+0.90$ & 0.005 & $+0.89$ & 0.007 & $+0.77$ & 0.072 \\
$F_1$ (all nodes) & $+0.76$ & 0.045 & $+0.64$ & 0.119 & $+0.53$ & 0.276 \\
\bottomrule
\end{tabular}
\end{center}

\begin{center}
\small
\begin{tabular}{lrrrr}
\toprule
Model & Jaccard & Recall & Precision (explicit) & F1 (all nodes) \\
\midrule
\texttt{minimax-m2.1}         & 0.96 & 0.98 & 0.99 & 0.99 \\
\texttt{gemma3:12b}           & 0.94 & 0.94 & 0.92 & 0.82 \\
\texttt{mistral-large-3:675b} & 0.80 & 0.84 & 0.83 & 0.65 \\
\texttt{ministral-3:8b}       & 0.77 & 0.94 & 0.91 & 0.79 \\
\texttt{minimax-m2.7}         & 0.70 & 1.00 & 0.89 & 0.83 \\
\texttt{qwen3-coder-next}     & 0.58 & 0.89 & 0.81 & 0.75 \\
\texttt{gpt-oss:120b}         & 0.39 & 0.60 & 0.59 & 0.56 \\
\bottomrule
\end{tabular}
\end{center}

\section{Stabilisation in full, and the recovery-campaign completion counts}
\label{app:rq5}

\paragraph{Recovery-campaign panel completion.}
The crossed campaign design calls for $11 \times 19 \times 3 = 627$ outputs. 557 persisted. Of the 209 model-prompt cells, 156 have three runs, 40 have two, nine have one and four are empty; the 196 with at least two runs are what \S\ref{sec:results} analyses. Of the 30 cells scoring exactly zero on exact-id Jaccard, 19 have three runs and 11 have two. The 70 missing outputs are not missing at random: a model that fails on a prompt is likelier to fail on it again, so the analysed panel is mildly biased toward models that complete.

\subsection{How many reruns a consumer needs}

Two specification errors surfaced here under audit, and both are instructive enough that we report this question mainly as a cautionary result.

\paragraph{The signal is a proxy.}
We do not measure a downstream tool. We measure the IR's \emph{node count} across runs, as a model-agnostic stand-in for any quantity a consumer might derive from the IR. No linter finding, coverage value, or mutation score is computed. Conclusions here transfer to a real tool only insofar as its output tracks node count, which we have not established.

\paragraph{The criterion is satisfied by construction at its endpoint.}
We defined $r$ as the smallest number of runs whose rolling mean stays within a tolerance of the long-run mean thereafter. With $R$ runs the long-run mean \emph{is} the mean of all $R$, so at $r=R$ the criterion compares a quantity with itself and always passes; "did not stabilise" is unreachable. At $R$ between 2 and 5 that accounts for 48 of 98 cells. Right-censoring $r=R$ is the minimum fix.

\paragraph{The tolerance was documented as relative and implemented as absolute.}
The flag is described as relative to the long-run mean; the code compares an absolute difference of 0.05 \emph{nodes}. On trajectories whose long-run means range from 5.3 to 18.0 nodes that is the difference between near-exact equality and roughly one node of slack. Cells stabilising before their endpoint go from 50 to 80 of 98, censored from 48 (49\%) to 18 (18\%); \texttt{qwen3-coder-next} 3/20 against 14/20 (Appendix~\ref{app:tolerance}). Under one reading half the study never demonstrably converges; under the other four fifths do. Nothing in the data distinguishes them; one undocumented word in a help string does.

\paragraph{What we claim.}
Only this: among cells that stabilise strictly before their endpoint, the median is $r=1$ under both readings, so where averaging helps it tends to help immediately. Any statement of the form "$r$ reruns suffice" is not supportable from this design.

\section{Failure attribution: harness budget versus model}
\label{app:attribution}
Each model probed twice on the same prompt, under a deliberately tight output budget (1500 tokens) and with room (8000). The tight condition is an artificial counterfactual, not a harness setting: the historical harness sent no token cap at all. Regenerated by \texttt{scripts/failure\_attribution.py}.

\begin{center}
\small
\begin{tabular}{lccccrr}
\toprule
& \multicolumn{2}{c}{Tight budget} & \multicolumn{2}{c}{With room} & & \\
\cmidrule(lr){2-3}\cmidrule(lr){4-5}
Model & finish & JSON & finish & JSON & lat (s) & reasoning \\
\midrule
\texttt{gpt-oss:20b} & length & invalid & stop & valid & 15.0 & 1,478 \\
\texttt{gpt-oss:120b} & length & invalid & stop & valid & 17.2 & 1,888 \\
\texttt{mistral-large-3:675b} & stop & valid & stop & valid & 13.5 & 0 \\
\texttt{minimax-m2.7} & length & invalid & stop & valid & 46.7 & 10,755 \\
\texttt{gemma4:31b} & stop & valid & stop & valid & 3.7 & 0 \\
\texttt{qwen3.5:397b} & length & invalid & stop & valid & 50.5 & 11,570 \\
\texttt{glm-5.2} & stop & valid & stop & valid & 12.3 & 4,372 \\
\texttt{deepseek-v4-pro:0813} & length & invalid & stop & valid & 33.8 & 18,255 \\
\texttt{nemotron-3-nano:30b} & length & invalid & stop & valid & 16.9 & 5,105 \\
\texttt{nemotron-3-super} & length & invalid & stop & valid & 13.0 & 1,209 \\
\texttt{nemotron-3-ultra} & length & invalid & stop & valid & 77.7 & 4,707 \\
\bottomrule
\end{tabular}
\end{center}

\section{Merge-rule sensitivity, full table}
\label{app:merge}
Per-model Jaccard / ID-stability / node-set-perfect count under each precedence rule.

\begin{center}
\small
\begin{tabular}{lcc}
\toprule
Model & Higher-budget-first & Earliest-first \\
\midrule
\texttt{gemma3:12b}           & 0.944 / 0.921 / 15 & 0.927 / 0.882 / 14 \\
\texttt{ministral-3:8b}       & 0.769 / 0.707 / 6  & 0.686 / 0.531 / 2  \\
\texttt{gpt-oss:20b}          & 0.757 / 0.687 / 4  & 0.680 / 0.495 / 0  \\
\texttt{qwen3-coder-next}     & 0.578 / 0.470 / 0  & 0.647 / 0.534 / 0  \\
\midrule
Node-set-perfect, all cells   & 35/127 (27.6\%)    & 26/127 (20.5\%)    \\
\bottomrule
\end{tabular}
\end{center}

\section{Stabilisation under both tolerance readings}
\label{app:tolerance}

\begin{center}
\small
\begin{tabular}{lrr}
\toprule
& Absolute (0.05 nodes) & Relative (5\%) \\
\midrule
Cells stabilising before endpoint & 50 & 80 \\
Cells right-censored              & 48 (49\%) & 18 (18\%) \\
\texttt{qwen3-coder-next}         & 3/20 & 14/20 \\
\texttt{mistral-large-3:675b}     & 12/19 & 19/19 \\
\bottomrule
\end{tabular}
\end{center}

\bibliographystyle{plainnat}
\bibliography{references}

\end{document}